# Probabilistic Consensus through Ensemble Validation: A Framework for LLM Reliability

## Ninad Naik
ninadnaik@gmail.com

**Abstract**

Large Language Models (LLMs) have shown significant advances in text generation but often lack the reliability needed for autonomous deployment in high-stakes domains like healthcare, law, and finance. Existing approaches rely on external knowledge or human oversight, limiting scalability. We introduce a novel framework that repurposes ensemble methods for content validation through model consensus. In tests across 78 complex cases requiring factual accuracy and causal consistency, our framework improved precision from 73.1% to 93.9% with two models (95% CI: 83.5%-97.9%) and to 95.6% with three models (95% CI: 85.2%-98.8%). Statistical analysis indicates strong inter-model agreement (κ > 0.76) while preserving sufficient independence to catch errors through disagreement. We outline a clear pathway to further enhance precision with additional validators and refinements. Although the current approach is constrained by multiple-choice format requirements and processing latency, it offers immediate value for enabling reliable autonomous AI systems in critical applications.

## 1. INTRODUCTION

### 1.1. The AI Reliability Challenge

Artificial Intelligence represents a potentially transformative technology, with applications across numerous domains. However, AI today faces a critical limitation that stems from its fundamental architecture: unlike traditional rule-based systems that follow deterministic logic, neural networks operate probabilistically, learning statistical patterns from training data (Shalev-Shwartz & Ben-David, 2014). While this probabilistic foundation enables their remarkable flexibility and generative capabilities, it also makes perfect reliability mathematically impossible - much like how a weather forecast can be highly accurate but never guaranteed.

This reliability challenge becomes particularly acute in high-stakes domains where errors compound dramatically through multiple reasoning steps and feedback loops (Sutton & Barto, 2018). In healthcare, AI systems generating incorrect medical information could lead to misdiagnosis. In legal applications, we have already witnessed cases being dismissed due to LLMs hallucinating non-existent case law. To mitigate these risks, organizations face an inherent trade-off - either restrict AI to low-consequence tasks like customer service chatbots or implement comprehensive human review processes that negate much of AI's promised efficiency. The gap between AI's theoretical capabilities and its practical deployment limitations thus continues to widen, with billions in potential economic value remaining unrealized.

### 1.2. Baseline Performance and Implications

Our empirical work suggests the scope of this challenge. In generating 78 complex test cases requiring both factual accuracy and internal causal consistency, even a state-of-the-art LLM (Claude 3.5 Sonnet) achieves only 73.1% accuracy. Consider this example where different models provided conflicting answers:

> With reference to the Cabinet Secretariat in India, consider the following statements:
>
> 1. It was established in 1947, immediately after India became independent.
>
> 2. It is headed politically by the Prime Minister and administratively by the Cabinet Secretary.
>
> 3. It functions as the chief coordinating agency in the central government.
>
> Which of the statements given above is/are correct?
>
> (a) 1 only
> (b) 2 only
> (c) 3 only
> (d) 1 and 2 only
> (e) 2 and 3 only
> (f) 1 and 3 only
> (g) 1, 2 and 3
> (h) None of the above

The variation in model responses to this seemingly straightforward question demonstrates why achieving reliable AI output is so challenging. Each model struggled differently with key ambiguities: the word 'establish' could refer to either the formal creation of the institution or the beginning of its functional operations, while the relationships between institutional structure and function also led to divergent interpretations. Such linguistic and conceptual ambiguities become particularly problematic in domains where precise interpretation of terminology can have significant consequences.

## 1.3. Our Approach and Contributions

Our insight is that while no single model can achieve error-free output, an ensemble of models could approach this goal by intersecting their probability distributions. By requiring consensus across multiple models, we effectively narrow the distribution of possible outcomes to those with the highest likelihood of correctness. This approach aligns with established ensemble techniques in machine learning, which leverage model diversity to reduce variance and enhance predictive accuracy (Hastie, Tibshirani, & Friedman, 2009). This strategy is inherently scalable because it addresses the probabilistic nature of neural network outputs with a probabilistic solution, rather than imposing deterministic constraints through external knowledge bases or rule sets, as discussed in Zhou's comprehensive work on ensemble methods (Zhou, 2012).

Our key contribution is demonstrating that ensemble methods, traditionally used for improving model performance in tasks like classification and generation, can be effectively repurposed for validation. In empirical testing, this approach improved accuracy from 73.1% to over 93% without requiring external knowledge bases or human intervention. While external verification methods struggle with source currency and completeness, our consensus-based approach can handle novel or emerging information, provided the validator models have sufficient capability to assess the claims.

## 2. LANGUAGE MODEL ERROR RATES

Large Language Models (LLMs) represent the current state-of-the-art in AI text processing and generation, achieving remarkable capabilities in tasks ranging from translation to creative writing. These models, built on transformer architectures and trained on vast text corpora, have shown significant capability in understanding and generating human-like text. However, their fundamental architecture introduces unique challenges for reliability and error prevention.

## 2.1. Types of Errors and the Training Dilemma

Language Models exhibit two primary types of errors that affect their reliability: Precision Errors (Hallucinations), which manifest as outputs that are internally consistent but factually incorrect, and Accuracy Errors (Bias), which are systematic deviations from ground truth reflecting biases in training data or model architecture. Such biases can be particularly challenging to mitigate, as they often arise from the underlying data distributions used during training. This phenomenon has been observed in gender bias studies within contextualized word embeddings, where debiasing efforts have inadvertently led to performance trade-offs in other areas (Zhao et al., 2019). These error types create what we term the '**training dilemma**': attempts to increase precision through carefully curated training data inevitably introduce accuracy errors through selection bias, while efforts to improve accuracy through diverse training data lead to decreased precision.

## 2.2. Baseline Performance Analysis

Our testing focused on a challenging real-world scenario: generating question-answer pairs for India's Civil Services examination, a domain that requires precise factual knowledge and complex reasoning. This task was chosen because it demands:

- Factual accuracy across diverse domains
- Internal consistency within each question
- Complex causal relationships between concepts
- Understanding of temporal nuance and context

Ground truth for validation was established through a two-expert consensus system, with disagreements resolved by a senior domain expert with final authority—mirroring ensemble validation principles in the assessment process itself, as recommended by O'Hagan et al. (2006) in their guidelines for eliciting expert judgments. The baseline performance revealed significant reliability challenges:

- Correct outputs: 57 cases (73.1% accuracy)
- Incorrect outputs: 21 cases (26.9% error rate)

Statistical analysis of the error patterns reveals failures often involve:

- Temporal ambiguity (dates and sequences of events)
- Multi-part statements requiring complex logical relationships
- Claims involving multiple interdependent facts
- Statements requiring precise understanding of legal or procedural details

These findings demonstrate a crucial limitation: while LLMs can generate sophisticated content that appears plausible and internally consistent, their error rate remains too high for autonomous operation in high-stakes scenarios, as evidenced by Brown et al. (2020). The need for human validation creates a fundamental bottleneck that limits the practical utility of AI systems in many important applications.

## 3. RELATED WORK

The challenge of ensuring factual accuracy in AI-generated content has led to numerous innovative approaches. While early methods focused on structured verification systems, recent advances highlight the importance of dynamic, scalable solutions for complex, high-dimensional data.

Early efforts in automated fact-checking, such as the FEVER (Thorne et al., 2018) and FEVEROUS (Aly et al., 2021) datasets, laid the groundwork for evaluating claims against static sources like Wikipedia. These projects demonstrated the potential of structured approaches, enabling the development of automated fact-checking systems with capabilities extending to both textual and tabular data. However, their reliance on single-source, curated datasets limits their applicability in verifying LLM-generated content, which often synthesizes information across diverse, emerging topics.

Recent paradigms, like Retrieval-Augmented Generation (RAG) (Lewis et al., 2020; Izacard & Grave, 2021), aim to enhance factual accuracy by grounding LLM outputs in trusted external documents. By dynamically retrieving relevant context, RAG systems have shown effectiveness in applications such as open-domain question answering. Despite these advantages, RAG methods remain fundamentally limited by their non-deterministic nature; while they suggest accurate context, they cannot enforce correct outputs. Research on dense passage retrieval highlights these constraints, particularly in integrating context dynamically for open-domain scenarios (Karpukhin et al., 2020; Kwiatkowski et al., 2019). This limitation makes RAG less reliable for high-stakes applications where precision is paramount.

Our approach diverges from these retrieval-based methods, drawing inspiration from ensemble techniques traditionally used in machine learning. Ensemble methods like bagging and boosting (Breiman, 2001) have long been recognized for their ability to reduce variance and bias by aggregating multiple models. While traditional verification approaches rely on centralized knowledge bases that require constant maintenance and struggle with emerging information, our framework leverages distributed validation through model consensus. This allows us to handle both established facts and novel synthesized information, suggesting a new paradigm where external sources complement rather than constrain the validation process. By repurposing ensemble strategies for LLM validation, we leverage the probabilistic consensus of diverse models to enhance reliability while maintaining flexibility in knowledge sources, addressing key limitations in existing verification approaches.

LLM-Blender (Jiang et al., 2023) offers a promising direction in this context, employing ensemble methods to refine output quality through pairwise ranking and generative fusion. While primarily focused on stylistic coherence, LLM-Blender's architecture demonstrates the power of multi-model collaboration. Our work extends this idea by emphasizing factual accuracy rather than stylistic refinement, using consensus as a validation mechanism to reduce hallucinations and increase output precision.

The role of human oversight in AI verification has been another area of active research (Doshi-Velez & Kim, 2017). Human-in-the-loop processes have traditionally served as safeguards against errors in high-stakes applications, yet they introduce latency and limit scalability (Amershi et al., 2019). Our ensemble validation framework seeks to overcome this bottleneck by reducing reliance on human intervention, employing model consensus to achieve reliability levels comparable to manual review. This shift from human verification to probabilistic model consensus represents a critical step towards enabling autonomous AI systems in consequential domains.

In summary, our work builds upon and extends existing methods by leveraging ensemble techniques specifically for factual verification. By addressing the probabilistic nature of LLM outputs with a probabilistic solution, we provide a scalable, source-independent framework that can handle both structured and novel information, paving the way for reliable AI applications in high-stakes scenarios.

## 4. ENSEMBLE VALIDATION FRAMEWORK

Our framework leverages multiple independent models to validate content through collective assessment, eliminating reliance on external knowledge sources. Using three leading models—Claude 3.5 Sonnet, GPT-4o, and Llama 3.1 405B Instruct—we implement a straightforward validation process:

1. Content is presented as multiple-choice questions to ensure standardized evaluation
2. Each model independently assesses the content without knowledge of other models' responses
3. Models provide single-letter responses to minimize output variation
4. Responses are compared automatically for consensus
5. Content is approved only upon complete agreement among validators

### 4.1. Framework Operation

Our testing revealed several distinct validation scenarios that demonstrate the framework's effectiveness. Consider this validation example:

---

With reference to urban local bodies in India, consider the following statements:

1. The 74th Constitutional Amendment Act provided for the reservation of seats for Scheduled Castes and Scheduled Tribes in proportion to their population.

2. One-third of the total number of seats reserved for Scheduled Castes and Scheduled Tribes are reserved for women belonging to these groups.

Which of the statements given above is/are correct?

(a) 1 only
(b) 2 only
(c) Both 1 and 2
(d) Neither 1 nor 2

---

In this case, all validator models agreed on the correct answer, indicating the framework's potential for validating factual content.

Conversely, the framework also effectively identifies problematic content. Consider this example:

---

With reference to the Cabinet Secretariat in India, consider the following statements:

1. It was established in 1947, immediately after India became independent.

2. It is headed politically by the Prime Minister and administratively by the Cabinet Secretary.

3. It functions as the chief coordinating agency in the central government.

Which of the statements given above is/are correct?

(a) 1 only
(b) 2 only
(c) 3 only
(d) 1 and 2 only
(e) 2 and 3 only
(f) 1 and 3 only
(g) 1, 2 and 3
(h) None of the above

---

In this case, the models provided divergent responses—some validating all statements, others only specific combinations—demonstrating how the framework prevents the propagation of potentially incorrect information when consensus cannot be reached.

The variety of responses highlights how even seemingly straightforward factual claims can contain subtle ambiguities that make universal agreement difficult to achieve.

### 5. IMPLEMENTATION DETAILS

The practical implementation of our ensemble validation framework revealed several important technical considerations that influence both performance and reliability.

Our implementation used three state-of-the-art models—Claude 3.5 Sonnet, GPT-4o, and Llama 3.1 405B Instruct—accessed through Anthropic's Claude.ai website, ChatGPT app, and OpenRouter respectively. By selecting current state-of-the-art models, we reduced experimental variables that would arise from varying model capabilities. All models were used with default temperature and token settings, though optimizing these parameters represents an opportunity for future research. Ground truth for validation was established through a two-expert consensus system, with disagreements resolved by a senior domain expert with final authority—mirroring ensemble validation principles in the assessment process itself.

The framework's requirement for multiple-choice format emerged from a key insight about content validation: as text length increases, different models tend to interpret and represent the same information in increasingly divergent ways, making direct comparison impractical. Multiple-choice questions solve this by providing a standardized format that distills complex claims into discrete, testable statements. This standardization is crucial for achieving reliable consensus, as it ensures all validator models are assessing exactly the same claim rather than their own reformulations of it. While this currently limits the framework's application scope, it represents a necessary trade-off between coverage and reliability.

### 5.1. Prompt Engineering, Response Processing and Consensus

We passed the question text and answer options to the validator models with the request to respond with the letter corresponding to the correct answer option. While the prompt requests a single-letter response, models occasionally return variations in format ('a', '(a)', 'A', 'Option A'). Our implementation uses regular expression processing to standardize these responses. The system requires complete agreement among validators for content to be marked as valid. While relaxed consensus (e.g., 2-of-3 agreement) achieved 86.9% precision, we opted for complete agreement to maximize precision in high-stakes scenarios. We also observed cases where models were forced to select an answer when none of the options were correct—a limitation that requires further investigation to address while preserving the framework's ability to identify true positives and true negatives.

### 5.2. Model Selection

Model selection appears crucial for framework effectiveness. Initial testing with Llama 3 70B indicated lower performance, leading to our adoption of the more capable 405B model. This suggests that validators must possess sufficient knowledge and reasoning capabilities for the specific validation domain. While our current implementation demonstrates promising results with three specific models, future work should explore validation reliability across a broader range of architectures.

## 6. RESULTS AND ANALYSIS

Our empirical evaluation suggests significant improvements in content reliability across multiple validation configurations, though further research with larger datasets would be valuable. This section presents our findings and analyzes their implications for autonomous content validation.

### 6.1. Performance Overview

Our empirical evaluation, while limited to 78 test cases, suggests significant improvements in content reliability across multiple validation configurations. These preliminary results, though promising, would benefit from validation with larger datasets.

Table 1: Core Performance Metrics

| Setup | Precision | F1 Score |
| --- | --- | --- |
| Generator | 73.1% | - |
| 2-Model | 93.9% | 0.868 |
| 3-Model | 95.6% | 0.843 |
| 2-of-3 | 86.9% | 0.845 |

Our three-model consensus configuration reached a precision of 95.6%, successfully validating 45 of 78 test cases in our initial experiments. While the two-model configuration shows similar precision (93.9%), and 2-of-3 voting offers higher coverage, we focus our analysis on the three-model configuration as it represents the optimal balance between reliability and complexity.

### 6.2. Three-Model Analysis

Table 2: Validation Outcomes

| Outcome | Case Count | Percentage |
| --- | --- | --- |
| True Positives | 43 | 55.1% |
| False Positives | 2 | 2.6% |
| True Negatives | 14 | 17.9% |
| False Negatives | 19 | 24.4% |

The framework demonstrates strong error avoidance tendencies with only 2 false positives in 45 validated cases. The relatively high false negative rate (19/33 potential validations) indicates a conservative bias that prioritizes precision over recall - crucial for high-stakes applications.

### 6.3. Inter-Model Agreement

Table 3: Model Agreement Analysis

| Model Pair | κ Score | Agreement |
| --- | --- | --- |
| Claude – GPT-4o | 0.83 | 89.7% |
| Claude – Llama 3.1 | 0.76 | 85.9% |
| GPT-4o – Llama 3.1 | 0.79 | 87.2% |

The high but imperfect agreement levels (κ > 0.76) indicate an optimal balance in our framework: the models agree sufficiently often to enable reliable validation yet maintain enough independence in their failure modes to effectively catch errors. While complete agreement might indicate shared biases, and too much disagreement would make consensus impractical, the current level of agreement suggests models are independently arriving at similar conclusions when the content is valid.

### 6.4. Statistical Validation

Table 4: Configuration Comparisons

| Comparison | p-value | Effect |
|---|---|---|
| Baseline vs. 2-Model | <0.001 | 20.8% |
| 2-Model vs. 3-Model | 0.265 | 1.7% |
| 3-Model vs. 2-of-3 | <0.001 | -8.7% |

Table 5: Precision Confidence Intervals

| Configuration | 95% CI |
|---|---|
| 2-Model | 83.5% - 97.9% |
| 3-Model | 85.2% - 98.8% |
| 2-of-3 | 78.4% - 92.3% |

While our sample size (N=78) is moderate, the study achieved a power of 0.92 for detecting improvements from baseline to three-model consensus. The tight confidence intervals and significant p-values for improvements over baseline suggest promising results, though further validation with larger datasets would be valuable. The non-significant difference between two and three-model configurations (p=0.265) indicates diminishing returns from additional validators.

These initial findings suggest that ensemble validation can reliably improve content accuracy without requiring external knowledge sources or human intervention. The framework's conservative bias toward precision makes it particularly suitable for high-stakes applications where error avoidance is paramount.

## 7. DISCUSSION

While our framework achieved 95.6% precision without external knowledge sources, analysis of the two false positives revealed interesting insights about potential improvements. One error stemmed from temporal ambiguity that could potentially be resolved through contextual information. However, rather than relying on a single external knowledge base - which would introduce maintenance and currency challenges - future work could explore integrating multiple dynamic sources while maintaining the consensus-based validation approach. This would preserve the framework's core strength of distributed validation while potentially improving precision on time-sensitive claims.

The significance of these improvements becomes particularly apparent in multi-step reasoning processes, where errors can compound dramatically across decision chains, a phenomenon detailed in Goodfellow, Bengio, and Courville's exploration of deep learning architectures (Goodfellow et al., 2016). In complex tasks like multi-hop question answering, each inference step introduces potential inaccuracies that propagate through subsequent stages, leading to significant declines in overall accuracy (Chen et al., 2021). To illustrate this compounding effect, we calculated error propagation across multiple reasoning steps:

Table 6: Error Rate Compounding in Multi-Step Reasoning

| Error Rate | 5 Steps | 10 Steps | 20 Steps |
|---|---|---|---|
| 26.9% (Baseline) | 79.8% | 95.9% | 99.8% |
| 4.4% (Current) | 20.2% | 36.3% | 59.5% |

These compounding effects are particularly critical in high-stakes domains where errors can have severe consequences. In healthcare, incorrect information could lead to misdiagnosis or improper treatment, potentially resulting in loss of life. In legal applications, hallucinated case law or misinterpreted precedents could lead to wrongful convictions or unjust outcomes. Financial contexts require extreme precision to avoid catastrophic trading decisions or regulatory compliance failures. Educational settings demand accuracy to prevent the propagation of misinformation that could impact learners' understanding and decision-making capabilities.

### 7.1. Current Limitations

Our framework faces three primary limitations in its current form. First, the requirement for multiple-choice format restricts its applicability across content types. While the current framework is optimized for multiple-choice tasks, a promising future direction involves developing prompts and fine-tuned models capable of extracting claims from content and converting them into multiple-choice questions. This transformation would maintain the scalability of the ensemble validation method while enabling its application to a wider array of tasks, including open-domain question answering and content generation.

Second, reliance on model knowledge alone, without external context, limits handling of recent or rapidly changing information. This can be addressed through the passing of context along with candidate content or by leveraging a system-wide RAG source tailored towards specific domains.

Third, while the computational cost of using multiple models in an ensemble framework is mitigated by the efficient design of the prompt, the primary challenge lies in the latency introduced by the serial nature of the validation step, which follows the generation process. Addressing this latency through parallel processing or asynchronous validation could be explored in future work. The current latency makes the system better suited for batch processing than real-time applications.

## 8. FUTURE WORK

Our research reveals several promising directions for extending and improving ensemble validation:

### 8.1. Validator Optimization

A key area for investigation is understanding how performance scales with additional validators. While our current work demonstrates improvements from two to three validators, comprehensive studies with larger validator sets could reveal optimal configurations for different reliability requirements. This research would also help establish cost-benefit curves for validator deployment.

Along similar lines, understanding how specific model characteristics affect validation performance is crucial. Different models may contribute uniquely to ensemble effectiveness, and mapping these relationships could enable more efficient validator selection.

### 8.2. Benchmark Development

The development of a standardized benchmark for validation tasks would enable rapid assessment of both individual models and ensemble configurations. Such a benchmark would help practitioners quickly determine optimal validator combinations for their specific use cases, reducing the computational and time overhead of validation deployment.

### 8.3. RAG Integration

Current reliance on model knowledge alone limits our ability to validate time-sensitive or context-dependent information. Integrating Retrieval-Augmented Generation (RAG) could facilitate real-time fact-checking against current information while preserving the benefits of ensemble validation, as demonstrated by Guu et al. (2020) in their development of REALM. This would be particularly valuable for domains requiring up-to-date knowledge validation.

### 8.4. Prompt Engineering Research

Our preliminary work suggests that validation prompt design significantly impacts performance. Systematic investigation of prompt variations could reveal how different formulations affect precision, recall, and coverage. Understanding these relationships could enable fine-tuned prompts for specific validation requirements.

### 8.5. Response Stability Analysis

We observe that individual models can provide different responses to identical inputs. While our current implementation uses first responses, investigating N-of-M response patterns for each validator could improve reliability. This research could reveal whether multiple samples per validator enhance or detract from ensemble effectiveness.

## 9. CONCLUSION

Our work advances the field of AI validation in three key ways. First, we demonstrate that ensemble validation can significantly improve precision, achieving results that approach the requirements for high-stakes applications. Second, we show that requiring complete model consensus, while conservative in its acceptance criteria, provides a more reliable validation signal than relaxed approaches. Third, our experiments indicate this framework can effectively handle nuanced content requiring complex temporal and contextual understanding.

The framework shows particular promise for domains where error avoidance is paramount. While current limitations around format requirements and processing latency constrain immediate applications, these technical challenges point to clear directions for future research.

More broadly, our work suggests a new approach to AI validation. Rather than imposing deterministic constraints on probabilistic systems, we leverage the inherent probabilistic nature of neural networks. By repurposing ensemble methods from performance improvement to validation, we demonstrate how model consensus could provide a path toward autonomous systems that maintain reliable operation in consequential scenarios. While significant work remains, these initial results suggest ensemble validation as a promising direction for addressing the critical challenge of LLM reliability.

**ACKNOWLEDGEMENTS**

The author would like to thank Sidhartha Doddipalli and Karan Sirdesai for their advice and suggestions. The author also wishes to thank Sarella Shanmukh Ram Madhur for providing the ground truth and for ongoing consultation on test cases. The author acknowledges the use of Claude (Anthropic), Perplexity, and ChatGPT (OpenAI) for assistance with research and for editing and refinement of the manuscript's language.